\documentclass[11pt,a4paper]{article}

\usepackage[british]{babel}

\usepackage[a4paper,top=2cm,bottom=2cm,left=2.5cm,right=2.5cm,marginparwidth=1.75cm]{geometry}

\usepackage[style=apa, backend=biber]{biblatex} 
\DeclareLanguageMapping{british}{british-apa} 
\DeclareFieldFormat[article]{volume}{\apanum{#1}} 

\usepackage{amsmath}
\usepackage{graphicx}
\usepackage[colorlinks=true, allcolors=blue]{hyperref}
\usepackage{hyperref}
\usepackage{orcidlink}
\usepackage[title]{appendix}
\usepackage{mathrsfs}
\usepackage{amsfonts}
\usepackage{booktabs} 
\usepackage{caption}  
\usepackage{threeparttable} 
\usepackage{algorithm}
\usepackage{algpseudocode}
\usepackage{listings}
\usepackage{enumitem}
\usepackage{chngcntr}
\usepackage{booktabs}
\usepackage{lipsum}
\usepackage{subcaption}
\usepackage{authblk}
\usepackage[T1]{fontenc}    
\usepackage{csquotes}       
\usepackage{diagbox}

\usepackage{amsmath,amssymb,amsfonts}

\usepackage[export]{adjustbox}
\usepackage{multirow}
\usepackage{makecell}

\usepackage{setspace}
\usepackage{titlesec}
\titleformat{\section} 
  {\normalfont\Large\bfseries}{\thesection.}{1em}{}
  
\usepackage{float}   
\usepackage{caption} 


\title{\protect\begin{tabular}{@{}c@{}}\protect\includegraphics[width=2cm,height=2cm,valign=c]{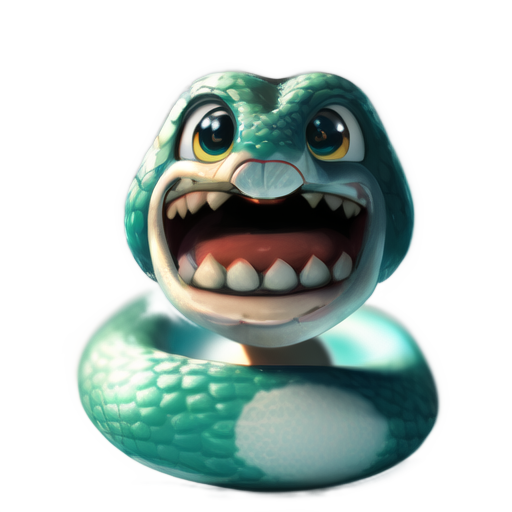}  \fontsize{15.5}{16}\selectfont T-Mamba: A unified framework with Long-Range Dependency in dual-domain for 2D \& 3D Tooth Segmentation\end{tabular}}

\author[1]{Jing Hao}
\author[2]{Yonghui Zhu}
\author[3]{Lei He}
\author[3]{Moyun Liu}
\author[1]{James Kit Hon Tsoi}
\author[1, *]{Kuo Feng Hung}
\affil[1]{The University of Hong Kong, Hong Kong SAR, China}
\affil[2]{Imperial College London, UK}
\affil[3]{Huazhong University of Science and Technology, China}
\affil[*]{Corresponding author: \texttt{hungkfg@hku.hk}}
\date{}  

\begin{document}

\maketitle

\begin{abstract}\label{abstract}
Tooth segmentation is a pivotal step in modern digital dentistry, essential for applications across orthodontic diagnosis and treatment planning. Despite its importance, this process is fraught with challenges due to the high noise and low contrast inherent in 2D and 3D tooth data. Both Convolutional Neural Networks (CNNs) and Transformers has shown promise in medical image segmentation, yet each method has limitations in handling long-range dependencies and computational complexity. To address this issue, this paper introduces T-Mamba, integrating frequency-based features and shared bi-positional encoding into vision mamba to address limitations in efficient global feature modeling. Besides, we design a gate selection unit to integrate two features in spatial domain and one feature in frequency domain adaptively. T-Mamba is the first work to introduce frequency-based features into vision mamba, and its flexibility allows it to process both 2D and 3D tooth data without the need for separate modules. Also, the TED3, a large-scale public tooth 2D dental X-ray dataset, has been presented in this paper. Extensive experiments demonstrate that T-Mamba achieves new SOTA results on a public tooth CBCT dataset and outperforms previous SOTA methods on TED3 dataset. The code and models are publicly available at \href{https://github.com/isbrycee/T-Mamba}{https://github.com/isbrycee/T-Mamba.}

\end{abstract}


\section{Introduction}\label{sec0}
The evolutionary key of modern digital dentistry lies in the acquisition and delineation of oral regions in different types of tooth data, including 2D panoramic radiography and 3D CBCT image. This technology finds several uses in oral and maxillofacial disciplines, including orthodontic diagnosis and treatment planning \cite{hung2020use, hung2022current, hung2023personalized}. Tooth segmentation is a crucial step in digital workflows, involving the extraction of a cluster of voxels from 3D digital models along with their intensity and density information, or the delineation of tooth regions from 2D grayscale panoramic radiographs. The process of tooth segmentation is critical in dental diagnostics, because it plays a key role in assessing conditions like impacted teeth, superordinate teeth, and missing dentition. Additionally, it aids in the review and evaluation of orthodontic treatment outcomes \cite{Teeth-U-Net}.
However, an accurate tooth segmentation process is challenging for several reasons. First of all, either the 2D panoramic radiography or 3D digital model is hard to observe due to the natural property of high noise and low contrast. In addition, the presence of metal fillings and prosthetic restorations in tooth image data introduces artifacts that induce distortions, significantly complicating the identification of teeth. Finally, the tooth data is typically acquired in natural occlusion, posing challenges in distinguishing between lower and upper teeth due to their similar densities.

Deep learning has been widely used in the field of medical image segmentation, and many researchers are dedicated to employing deep learning techniques for automated segmentation of tooth \cite{ToothNet, TSegNet, CTA-UNet, CoT-UNet++}. CNNs and Transformers architecture have displayed considerable promise in medical image segmentation due to their ability to learn complex image features, but both of them have limited ability to handle long-range dependencies because of inherent locality or computational complexity. The CNNs can only capture translational invariances and extract local features, and the Transformer is struggled by the heavy computational complexity when capturing global contextual information. Due to their complementary feature, many studies have explored incorporating Transformers into CNNs via hybrid network architectures.  Nonetheless, the obvious shortcoming of  Transformers is resource-intensive because the self-attention mechanism scales quadratically with the input size and poses challenges in terms of speed and memory usage with high-resolution biomedical images. While substantial effort has been dedicated to reducing the computational complexity of Transformers, they often come at the expense of sacrificing model accuracy \cite{SPSA, Edgenext}. Therefore, the efficient enhancement of long-range dependency in CNNs continues to be an unresolved issue.

Recently, motivated by the success of Mamba in language modeling, many studies have transferred this success from language to vision for the aim of achieving linear complexity without sacrificing global receptive fields \cite{VisionMamba, Vmamba}. However, we identify two limitations in utilizing vision mamba to assist CNNs for modeling long-range dependency. First and foremost, due to the inherent imaging principles of medical images such as CT and X-ray, these images possess natural attributes of high noise and low contrast from a visual perspective. For such images, frequency domain-based feature representations are more accurate, unique, and robust. However, CNNs and vision mamba models typically extract semantic features solely from the spatial domain, overlooking the rich frequency domain-based information. As indicated by \cite{ct_freq1}, high-frequency components capture texture details, whereas low-frequency components encode shape information. Therefore, integrating frequency domain features with spatial domain representations holds promise for enhancing the image feature extraction in medical images, consequently improving the accuracy of prediction. Secondly, CNNs can process 2D or 3D features directly while vision mamba structures is to handle 1D feature sequences. In the CNN and vision mamba hybrid architecture, the transformation of features by vision mamba inevitably leads to the loss of spatial position information. This spatial context is crucial, particularly in tasks requiring dense precise positional prediction. 

Inspired by these two limitations, we design a network named T-Mamba, which amalgamates our proposed Tim (Tooth Vision Mamba) block with DenseVNet \cite{DenseVNet} in multi-scale features, for tooth image segmentation. The Tim block exhibits three key strengths:
(1) It extracts image features in both frequency domain and spatial domain, thus we can derive more accurate, unique, and robust feature representations for medical images which are of high noise and low contrast. Based on the attributes of features at different scales, we have employed distinct bandpass filtering strategies tailored to each scale. 
(2) It contains a shared bi-position embeddings which are to compensate for the positional information lost during reshape operation. It is noteworthy that we employ a singular positional embedding within each feature-scale. This not only guarantees the preservation of spatial positions across feature maps of identical scales but reduces model parameters and computational burden.
(3) It includes a gate selection unit for integrating two features in spatial domain (both forward and backward directions) and one feature in frequency domain adaptively. The gate selection unit is data-dependent and can assign weights for three distinctive features according to input sequence tokens. 
Also, Our T-Mamba demonstrates the superior flexibility when processing 2D and 3D tooth data, without the need for separate, carefully designed modules for each type of data. When extracting feature for different types of data, changing the demention of CNN is adequate because the Tim block in the T-Mamba transforms the input data into 1D sequences for feature modeling. Therefore, T-Mamba provides a unified solution with long-range dependency for both 2D and 3D tooth segmentation.

In order to verify the effectiveness of T-Mamba, extensive experiments were conducted on both 3D tooth CBCT data and 2D tooth X-ray images. Due to the unaccessibility to a large-scale 2D tooth X-ray datasets, we build the TED3 dataset, a public Teeth large-scale 2D X-Ray Dental Dataset. This dataset is entirely based on the collection and integration of various publicly available panoramic dental X-ray datasets accessible online. It consists of 6225 labelled data and 14728 unlabelled data, which is the largest 2D tooth X-ray segmentation dataset to date.

Our T-Mamba outperforms previous state-of-the-art results in a large margin on a public 3D tooth CBCT dataset and our TED3 dataset. Besides, we also implement sufficient ablation studies to verified the effectiveness of three components in our proposed Tim block. To the best of our knowledge, T-Mamba is the pioneering effort to incorporate frequency domain features into the vision mamba framework. Our main contributions can be summarized as follows:
Our main contributions can be summarized as follows:
\begin{itemize}
    \item We propose T-Mamba, which incorporates our designed Tim block with DenseVNet for global and local visual context modeling for both 2D \& 3D Tooth segmentation. 
    \item The Tim block is the first work to extract more robust and unique representations for medical images which are of high noise and low contrast by introducing frequency-based features.
    \item We introduce a large-scale 2D tooth X-ray segmentation dataset, TED3, which consists of 6225 labelled data and 14728 unlabelled data.
    \item Without bells and whistles, T-Mamba achieves new SOTA results on the public Tooth CBCT dataset and outperforms previous SOTA methods in a large margin on our proposed TED3 dataset.
\end{itemize}

\section{Related Works}\label{sec1}
\textbf{Vision mamba and its variants.}
\cite{S4ND} is the first work applying SSMs into visual tasks and showing the potential that its performance may compete with ViT \cite{ViT}. \cite{Vmamba} traverse the spatial domain and convert any non-causal visual image into order patch sequences. \cite{VisionMamba} marks the image sequences with position embeddings and compresses the visual representation with bidirectional SSMs. \cite{Vivim} introduces the structured SSMs with spatiotemporal selective scan for medical video object segmentation task. \cite{U-mamba} which is a hybrid CNN-SSM architecture is proposed handle the long-range dependencies in biomedical image segmentation. In addition, there are some other works that leverage the SSMs into the medical image analysis field, point cloud analysis and vedio generation, such as \cite{Segmamba, Swin-umamba, P-Mamba, Vm-unet, pointmamba, Matten, liu2024rscama}.

\noindent
\textbf{3D Tooth CBCT Segmentation.}
Accurate and automatic depiction of individual teeth from CBCT images is a critical step to assist physicians in the diagnosis and treatment of oral diseases.
\cite{ToothNet} was the first to use neural networks to achieve automatic tooth segmentation and identification from CBCT images.
\cite{TSegNet} was designed for robust and efficient tooth segmentation on 3D dental scanned point cloud data.
\cite{fully_Jang} present a fully automated method of identifying and segmenting 3D individual teeth by hierarchical multi-step models.
\cite{CTA-UNet} designed a U-shaped network which combines the advantages of CNN and Transformer for dental CBCT images segmentation. 
\cite{CoT-UNet++} proposed a context-transformed architecture and obtained better performance in tooth CBCT segmentation.

\noindent
\textbf{2D Tooth X-Ray Segmentation.}
Accurate segmentation of teeth in panoramic dental X-rays is also another challenging task due to variations in tooth morphology and overlapping regions.
\cite{rubiu2023teeth} utilized a mask region-based convolutional neural network (Mask-RCNN \cite{mask_rcnn}) to segment tooth instances in Panoramic Dental X-ray. 
\cite{numbering_seg_2020} brought a thorough study on tooth segmentation and numbering on panoramic X-ray images by means of end-to-end deep neural networks.
\cite{Teeth-U-Net} proposed the Teeth U-Net by introducing context semantics and contrast enhancements upon U-Net. 
\cite{PaXNet} proposed an automatic diagnosis system to detect dental caries in Panoramic images, using ensemble transfer learning and capsule classifier.
\cite{lin2023lightweight_xray_seg} designed a novel lightweight neural network scheme using knowledge distillation for dental X-ray image segmentation for the purpose of deployment on edge devices. 
However, these neural networks are specifically designed for 2D tooth data and cannot be directly applied to 3D tooth CBCT segmentation tasks, thus lacking a certain degree of flexibility. Our framework, on the other hand, can perform segmentation tasks on both 2D and 3D data without the need for carefully designed network structures specific to different data modalities.

\section{Methodology}\label{sec2}
\subsection{Preliminaries for Mamba}
The advanced state space models (SSM), i.e., structured state space sequence models (S4) and Mamba, are a type of systems that maps a 1-D continuous function or sequence $x(t)\in\mathbb{R}\mapsto y(t)\in\mathbb{R}$ through a hidden state $h(t)\in\mathbb{R}^N$. Mathematically, these models are typically formulated as linear Ordinary Differential Equations (ODEs), as appears in (1):
\begin{equation}\label{eq1}
\begin{split}
   & h'(t)=\textbf{$\mathrm{A}$}h(t)+\textbf{$\mathrm{B}$}x(t), \\
   & y(t)=\textbf{$\mathrm{C}$}h(t),
\end{split}
\end{equation}
where the parameters include $\textbf{$\mathcal{A}$}\in\mathbb{R}^{N\times N}$ as the evolution parameter, and $\textbf{$\mathrm{B}$}\in\mathbb{R}^{N\times 1}$, $\textbf{$\mathrm{C}$}\in\mathbb{R}^{1\times N}$ as the projection parameters.

The SSM-based models, as continuous-time models, should be discretized when integrated into deep learning algorithms. This discretization transformation is crucial to align the model with the sample rate of the underlying signal embodied in the input data [1]. Given the input $x(t)\in\mathbb{R}^{L\times D}$, a sampled vector within the signal flow of length $L$ following [2], Equation (1) could be discretized as follows using the Zeroth-Order Hold (ZOH) rule:
\begin{equation}\label{eq2}
\begin{split}
   & h_t=\overline{\textbf{{A}}}h_{t-1}+\overline{\textbf{{B}}}x_{t} \\
   & y_t=\overline{\textbf{{C}}}h_{t} \\
   & \overline{\textbf{{A}}}=e^{\Delta \textbf{A}} \\
   & \overline{\textbf{{B}}}={\Delta \textbf{A}}^{-1}(e^{\Delta \textbf{A}}-\mathbf{I})\cdot \Delta \textbf{B} \\
   & \overline{\textbf{{C}}}=\textbf{{C}},
\end{split}
\end{equation}
where $\Delta \in\mathbb{R}^{D}$ is the timescale parameter. 

Eventually, the models compute output $y$ through a global convolution operation within a structured convolutional kernel $\overline{\textbf{{K}}}$:
\begin{equation}\label{eq3}
\begin{split}
   & \overline{\textbf{{K}}}=(\textbf{{C}}\overline{\textbf{{B}}}, \textbf{C} \overline{\textbf{{A}}} \overline{\textbf{B}}, \textbf{C} \overline{\textbf{{A}}}^{2} \overline{\textbf{B}},...,\textbf{C} \overline{\textbf{{A}}}^{L-1} \overline{\textbf{B}}), \\
   & y=\overline{\textbf{{K}}}\otimes y_t.
\end{split}
\end{equation}

\subsection{T-Mamba architecture}
\begin{figure}[ht]
\centering
\includegraphics[width=1.0\linewidth]{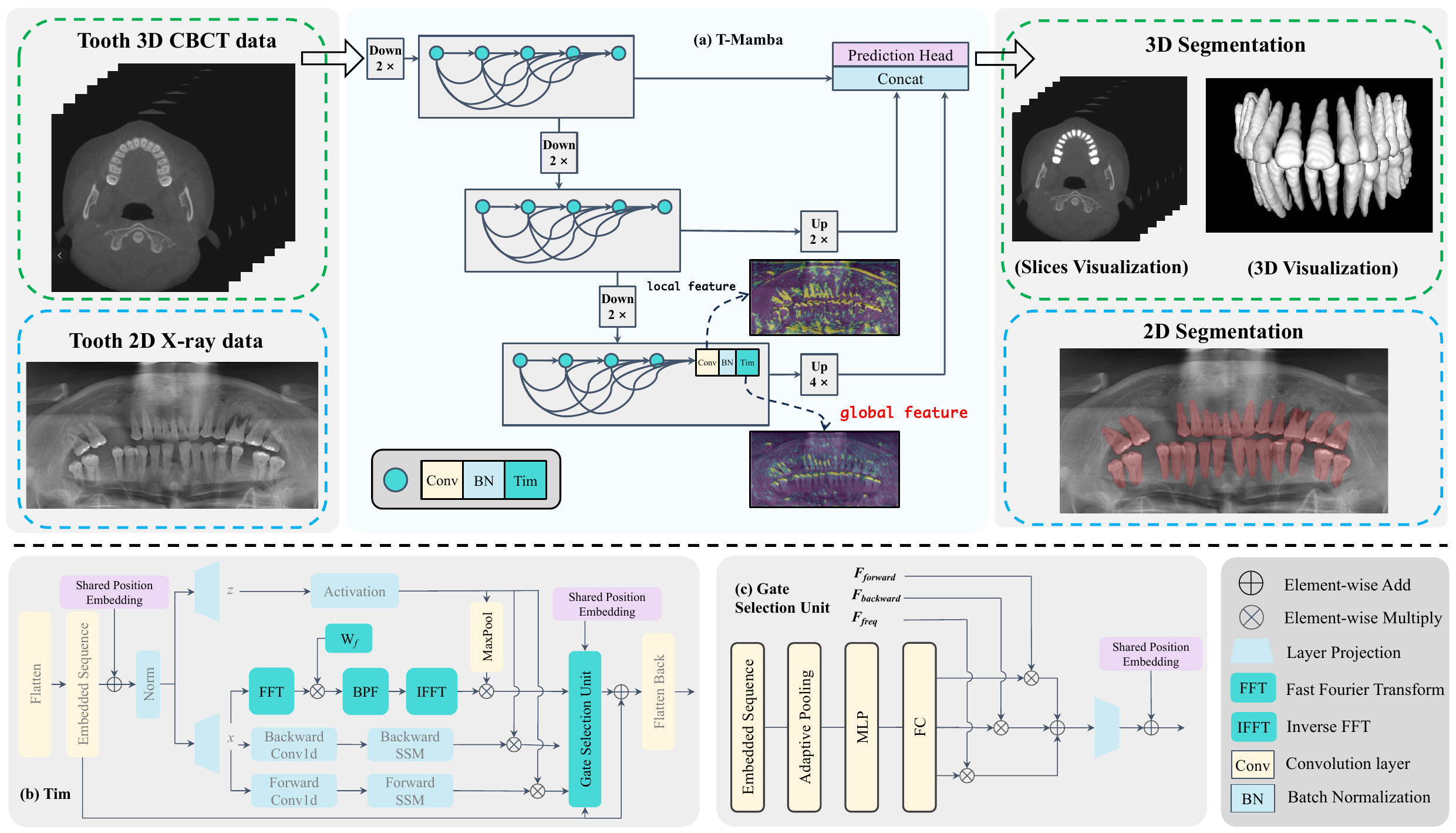}
\caption{\label{fig:framework}The framework of T-Mamba.}
\end{figure}

We enhance the global representation dependency of CNN by leveraging Mamba’s linear scaling advantage with three optimizations, and propose a unified and powerful framework, T-Mamba, for 2D \& 3D tooth semantic segmentation. The architecture of the T-Mamba is sketched in Fig. \ref{fig:framework}, which amalgamates our proposed Tim block with DenseVNet \cite{DenseVNet} in multi-scale features. 

The T-Mamba adopts the classical single-stage V-shaped architecture, and consists of three feature scales. To enlarge the receptive field of CNN and model the long-range dependency, we simply insert the Tim block after each CNN layer in DenseVNet. Different from previous works that performing global feature extraction directly in the pixel space of medical images \cite{Swin-umamba, U-mamba, Segmamba}, performing global modeling in a CNN-style feature space ,which is with inductive bias property, is less challenging to optimize model parameters and yields better performance \cite{Tfcns, tang2024htc, xin2023cnn}.

T-Mamba can capture both localized fine-grained feature and long-range dependencies in terms of spatial domain and frequency domain for 2D tooth X-ray image or 3D tooth CBCT data. This implementation is achieved by simply modifying the type of CNN used for extracting features in T-Mamba, such as 2D convolution or 3D convolution. For 2D tooth X-ray images, 2D convolutional operations are used, while for 3D tooth CBCT data, 3D CNN convolutional operations are employed. Note that our proposed Tim block does not require any modifications regardless of whether it is processing 2D or 3D data, as it transforms the input data into sequences for feature modeling. Therefore, our T-Mamba can flexibly perform feature modeling on both 2D and 3D data, without the need for meticulously designing different modules for various data modalities. After encoding the image features at three different scales, these features are concatenated together, and a lightweight prediction head is utilized to output the final segmentation results.

\subsection{Tim block}
The original Mamba block is devised for the 1-D sequence modeling, which is not compatible for vision tasks requiring spatial-aware understanding. The Vision Mamba \cite{VisionMamba} proposed Vim block which incorporates the bidirectional sequence modeling upon the Mamba block. However, directly applying this module to tooth modality data for feature modeling may yield suboptimal results due to the intrinsic differences between tooth data and natural images. Unlike natural images, tooth data typically comprises grayscale images with lower contrast and unclear object textures. Meanwhile, the Vim block neglects the spatial feature loss during the process of converting high-dimensional data into 1D sequential data.
Motivated by these problems, we further enhance the Vim block by introducing three components for better the feature modeling of tooth modality data: 1)  Frequency-based band pass filtering; 2)Shared bi-positional encoding compensation; 3) Gated selection unit. The Tooth Vision Mamba (Tim) block is shown in Fig. \ref{fig:framework}. 

\noindent
\textbf{Frequency-based band pass filtering.}
The Fourier domain plays a major role in extracting frequency-based analysis of image information and it’s evident that fine details and outlines could be captured in frequency domain even with poor quality X-ray \cite{xray_freq_1, Wavelet} and CT \cite{ct_freq1, UCFilTransNet}.
The convolutional layers have a strong texture inductive bias, and they tent to learn texture-based feature. Representing an object in terms of frequency may reduce the effect of texture bias since only the high frequencies are responsible for the texture information (like boundaries) and lower frequencies might be related to the shape and outline \cite{zhou2023xnet}.

Inspired by this, we evolve the structure of Vim by enhancing features extraction in frequency domain. Specifically, we firstly transform the sequence features X to the Fourier domain, then extract frequency features using learnable weight parameters $W_f$ and implement a bandpass filtering, and finally conduct the inverse Fourier transform to get the signal back. After that, the frequency feature is aggregated by the activated $Z'$ with maxpooling operation. Since image information varies at different scales of features, we have applied specific frequency domain filtering operations to different scales of features. For low-level features, which primarily contain the shape and outline of objects, we perform low-pass filtering operations to enhance the modeling of object contours and shape. For high-level features, high-pass filtering is implemented to enhance the modeling of texture information in images. For medium-level features, we apply band-pass filtering operations accordingly.

The whole process can be formulated as:
\begin{equation}
\label{eq4}
    \mathcal{F}_{freq} = IFFT(Bandpass(W_f(FFT(X))))*Maxpool(Z'), 
\end{equation}
\begin{equation}
\label{eq5}
\begin{split}
   & Bandpass = \begin{cases}  &  X*(|X| < S_{low}), X \in low-level\ features, \\  
   &  X*(S_{low} < |X| < S_{high}), X \in medium-level\ features,   \\  &  X*(|X| > S_{high}), X \in high-level\ features, \end{cases}
\end{split}
\end{equation}
where $S_{low}$, $S_{high}$ are the thresholds of bandpass filtering. In our experiments, we set $S_{low}$=0.1 and $S_{high}=0.9$. The low-level, medium-level, high-level indicates three different feature scales in our network, respectively. 

\noindent
\textbf{Shared Dual Positional Encoding Compensation.} Our T-Mamba network integrates the local feature extraction capabilities of convolutional layers with the SSMs for capturing the long-range dependency. 
Convolutional layers typically handle 2-D or 3-D feature maps, while our designed Tim block focuses on process 1-D sequences data. Consequently, reshaping high-dimensional features into 1-D feature tokens is essential. However, this process inevitably results in the loss of crucial positional information, which is vital for dense prediction tasks. To mitigate this, we employ position embeddings to compensate for the positional information lost during the reshape operation.

Specifically, given an input feature $\mathcal{I}$ with a shape of (B, C, H, W, D), we first flattened it to 1-D feature tokens with a shape of (B, L, C) where L = H × W × D and then add a learnable position embedding $\mathcal{F}_{pos}$ with a shape of (C, L) to feature tokens to retain positional information. 
\begin{equation}
\label{eq6}
    X = Flatten(\mathcal{I}) + \mathcal{F}_{pos}, 
\end{equation}
where X is sequence features which is fed to our Tim block. 

Regarding the output of Tim block, we also need to reshape the 1-D feature tokens into high-dimensional features for next convolutional operation. To further enhance the spatial information within the 1-D feature tokens, the positional embedding we used before is added to 1-D feature tokens again before reshaping them back into high-dimensional features. Note that the positional embedding at each feature scale remains consistent. This practice ensures spatial positions remain unchanged in feature maps of the same scale, simultaneously reducing model parameters and computational burden. Following \cite{attentionIsAllUNeed}, we initialize the position embedding via sinusoidal function:
\begin{equation}
\label{eq7}
\begin{split}
   & PE_{(pos, 2i)} = s i n(p o s/10000^{2i/d_{\mathrm{nodel}}}), \\
   & PE_{(pos, 2i+1)} = cos(p o s/10000^{2i/d_{\mathrm{nodel}}}),
\end{split}
\end{equation}
where pos is the position along with L and i is the index along with C. 

We posit that adding a shared bi-positional embedding to both the input and output of the Tim block significantly preserves the positional information of high-dimensional features. Through ablation experiments, we validate that employing a shared bi-positional encoding leads to higher performance compared to using a single positional embedding.

\noindent
\textbf{Gate Selection Unit.}
The evolved Vim block consists of two features in spatial domain (both forward and backward directions) and one feature in frequency domain. We devise a Gate Selection Unit with the objective of fusing these distinctive features adaptively. The input embedding sequence is firstly down-sampled to a fixed dimension such as 2048, and it is projected through MLP (Multi linear projection), along with a fully connection to predict three proportions corresponding to three features. After that, the $f_{fuse}$ is derived by weighted summation of three features and then projected by a linear layer, and Gate Selection Unit outputs the sum of the $f_{fuse}$ and the residual information. 
\begin{equation}
\label{eq8}
\begin{split}
   & S_{forward}, S_{backward}, S_{freq} = FC(MLP(Pooling(X))), \\
   & \mathcal{F}_{fuse} = \sum{(S_{i}*\mathcal{F}_{i}}), i \in {\{forward, backward, frequency\}}, \\
   & O_{gate} = \mathcal{F}_{fuse} + X.
\end{split}
\end{equation}

The Gate Selection Unit is data-dependent because the three weight coefficients are computed from source X, and these weight coefficients are then used to update different forms of features of X. Consequently, the Gate Selection Unit can adaptively adjust the combination of the three forms of features based on different inputs, thus obtaining a better feature representation.

\section{TED3 Dataset}
\label{sec3:ted3}

\begin{table*}[h]
\footnotesize
\centering
\caption{Composition of our proposed TED3 dataset.}
\label{tab: TED3 dataset}
\setlength{\tabcolsep}{3mm}{
    \begin{tabular}{c|ccccccc}
    \toprule
    Dataset & Whole  &Training Set & Test Set \\
    \midrule
    TED3-labelled & 6225  & 5000 & 1225  \\
    \midrule
    TED3-unlabelled & \multicolumn{3}{c}{14728} \\
    \bottomrule
    \end{tabular}}
\end{table*}

\begin{figure}[h]
    \centering
    \begin{subfigure}{0.48\textwidth}
        \centering
        \includegraphics[width=\textwidth]{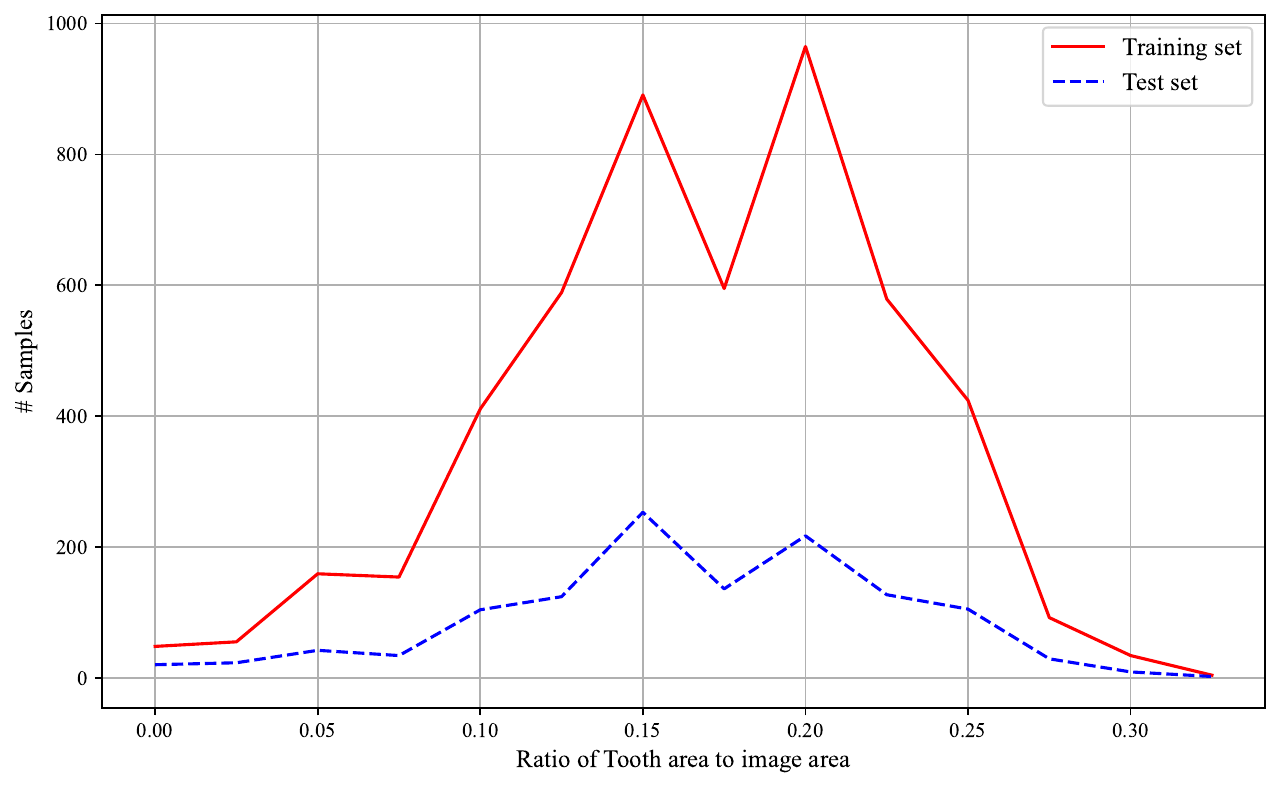}
        \caption{The mask ratio of Tooth area in TED3.}
        \label{fig: mask ratio of TED3}
    \end{subfigure}
    \begin{subfigure}{0.48\textwidth}
        \centering
        \includegraphics[width=\textwidth]{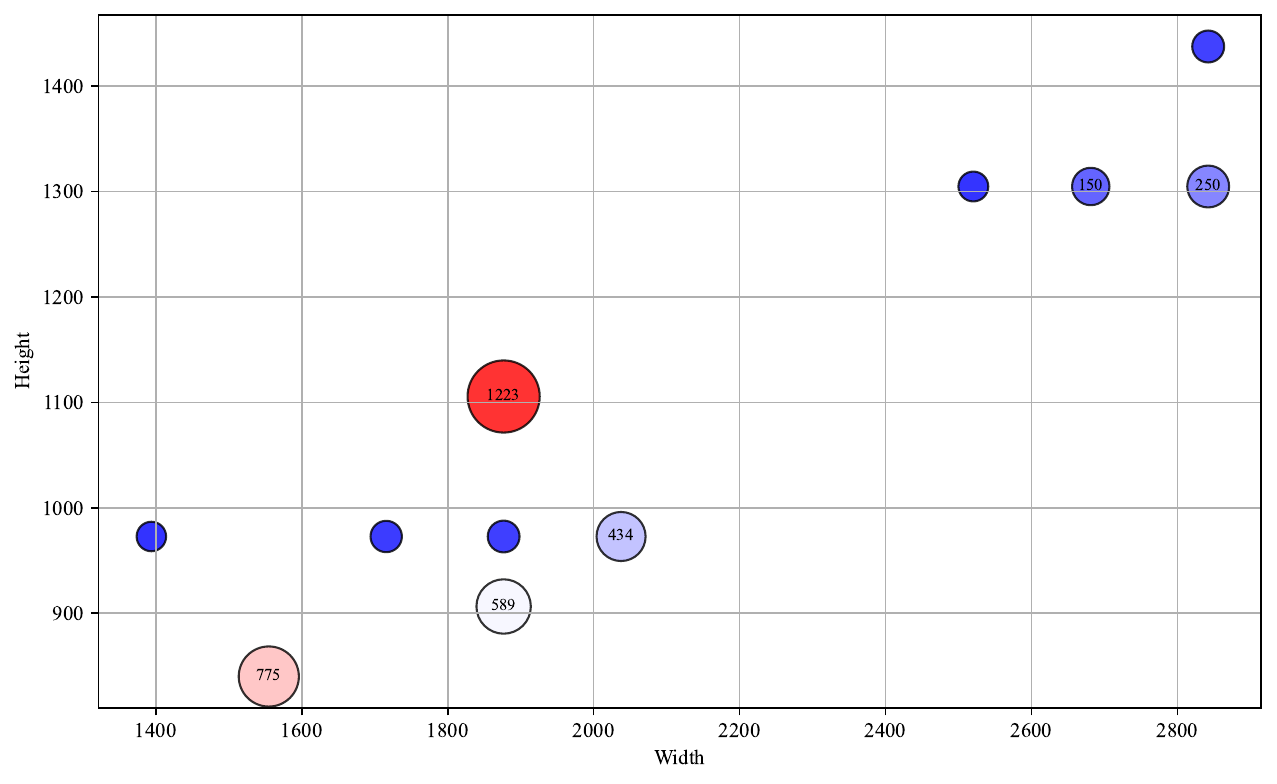}
        \caption{The image size distribution of TED3.}
        \label{fig: The image size distribution of TED3}
    \end{subfigure}
    \caption{The detailed statistics of TED3}
    \label{fig: The detailed statistics of TED3}
\end{figure}

Currently, most deep learning-based tooth segmentation algorithms for panoramic radiographs are designed and validated on private dataset because of the data protection, privacy concerns and annotation cost \cite{teeth_work1, teeth_work2, teeth_work3, teeth_work4}. This leads to the issue that most methods suffer from limited generalization ability due to the small training dataset size (typically less than 1000 images) and the lack of data diversity. Additionally, the majority of existing datasets are collected from the single center, resulting in the overfitting of center-specific characteristics during model training and thus impeding the generalization to data from different centers. These limitations hinder the advancement of deep learning in the field of dentistry.

To solve this issue, we build the TED3 dataset, a public \textbf{Te}eth large-scale 2\textbf{D} X-Ray \textbf{D}ental \textbf{D}ataset. This dataset is entirely based on the collection and integration of various publicly available panoramic dental X-ray datasets accessible online. 
Some small datasets originating from diverse health institutions for panoramic dental X-rays are available, so we systematically organize all these available public datasets, filter the identical data, and subsequently partition them into training and testing sets. 
We have compiled all publicly available panoramic dental X-ray datasets from the current online resources, with detailed information regarding their sources and dataset descriptions provided in the appendix.
Table \ref{tab: TED3 dataset} provides an overview of TED3. 

\begin{figure}[t]
\centering
\includegraphics[width=0.9\linewidth]{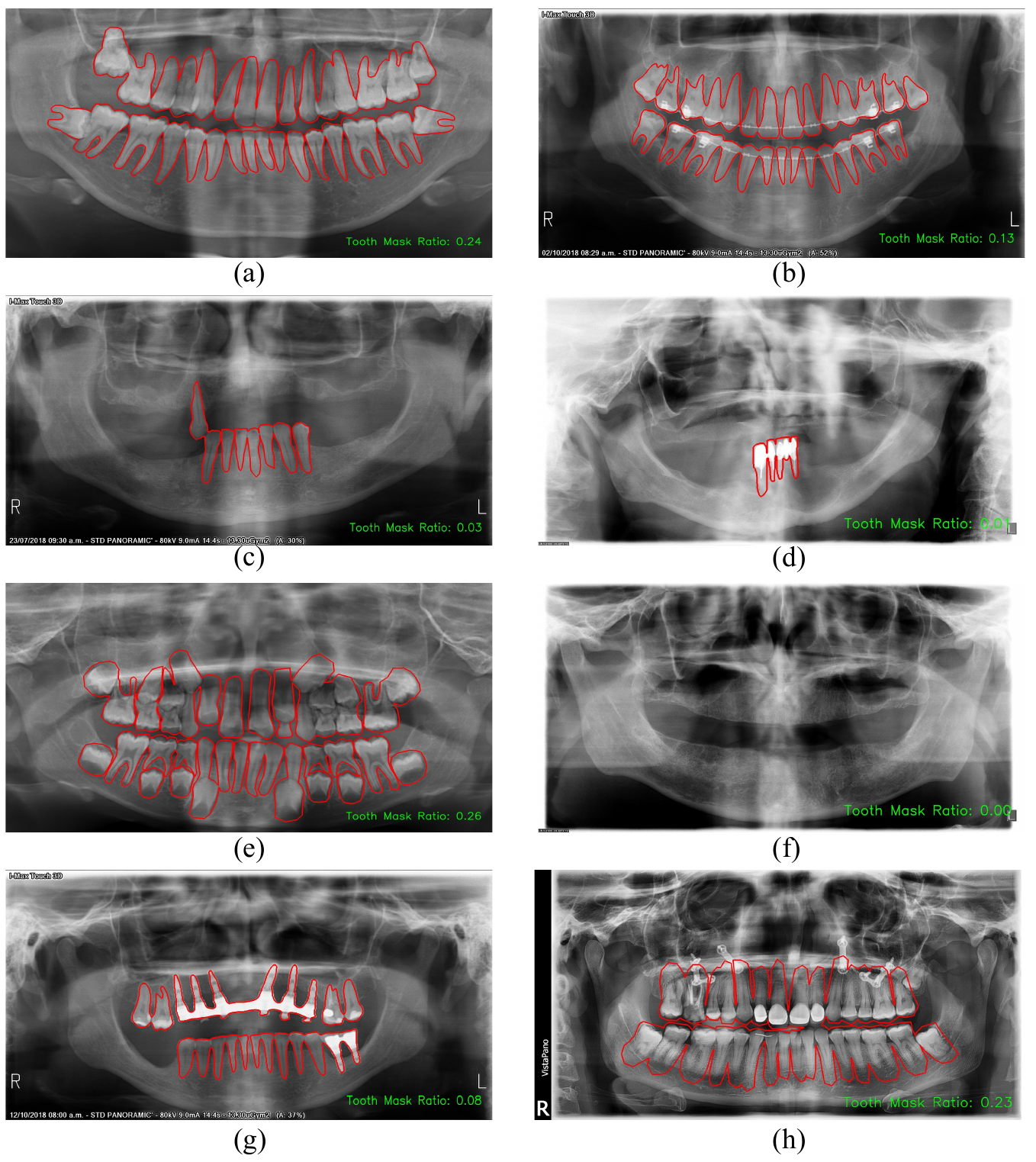}
\caption{The examples of TED3-labelled dataset in the test set. The Tooth Mask Ratio indicates the ratio of tooth area to image area. We only visualize the contours of masks for a better view.}
\label{fig: TED3_with_mask_crop}
\end{figure}

The TED3 consists of two parts: TED3-labelled data and TED3-unlabelled data. The TED3-labelled data includes 6225 images, with 5,000 training images and 1225 test images.
TED3-labelled is an amalgamation of multiple publicly available datasets from the internet, resulting in the diversity ranging from imaging devices, patient ages, and dental conditions. Therefore, the test set of TED3-labelled data enables comprehensive validation of the model's generalization ability, mitigating the risk of overfitting to a specific data distribution from a single health center. Figure \ref{fig: TED3_with_mask_crop} illustrates various samples in the test set, showcasing the diversity of test data distribution. 
Furthermore, we also provide the TED3-unlabelled data with 15069 unlabeled images. These data are primarily sourced from publicly available panoramic dental X-rays but lack binary masks. We believe that TED3-unlabeled data is beneficial for self-supervised learning and semi-supervised learning purposes, thereby further enhancing the accuracy of tooth segmentation models. 

The detailed statistics of TED3 are presented in Figure \ref{fig: The detailed statistics of TED3}. Figure \ref{fig: mask ratio of TED3} depicts the ratio of tooth area to image area for both the training and test sets, revealing a significant variation in the mask ratio of teeth ranging from 0 to above 0.3 (refer to Figure \ref{fig: TED3_with_mask_crop} for examples with different mask ratios of teeth). 
The TED3 dataset demonstrates remarkable diversity across its training and validation sets, as evidenced by variations in the mask ratio of teeth. This metric serves as a reflection of both the quantity and size of teeth captured within the images. Encompassing a wide array of conditions such as metal artifacts, missing teeth, misalignment issues, and cases without dental abnormalities, this dataset encapsulates a comprehensive spectrum of dental health scenarios observed across different age demographics, including children, adults, and the elderly.
Consequently, models trained on this dataset are highly likely to exhibit robust generalization capabilities and mitigate the risk of overfitting to single-center-specific characteristics. Figure \ref{fig: The image size distribution of TED3} illustrates the distribution of image sizes and quantities in the TED3-labeled dataset.


\section{Experiments and Discussion}\label{sec4}

We conduct extensive comparisons with a range of stage-of-the-art (SOTA) methods and numerous vision mamba variants, demonstrating the superiority of our T-Mamba and achieving new stage-of-the-art results in terms of 2D \& 3D tooth segmentation tasks. Besides, we also did sufficient ablation studies on Tim block for proving the effectiveness of each components. 

\subsection{Experiment Settings}
\subsubsection{Dataset}
\begin{figure}[ht]
\centering
\includegraphics[width=0.55\linewidth]{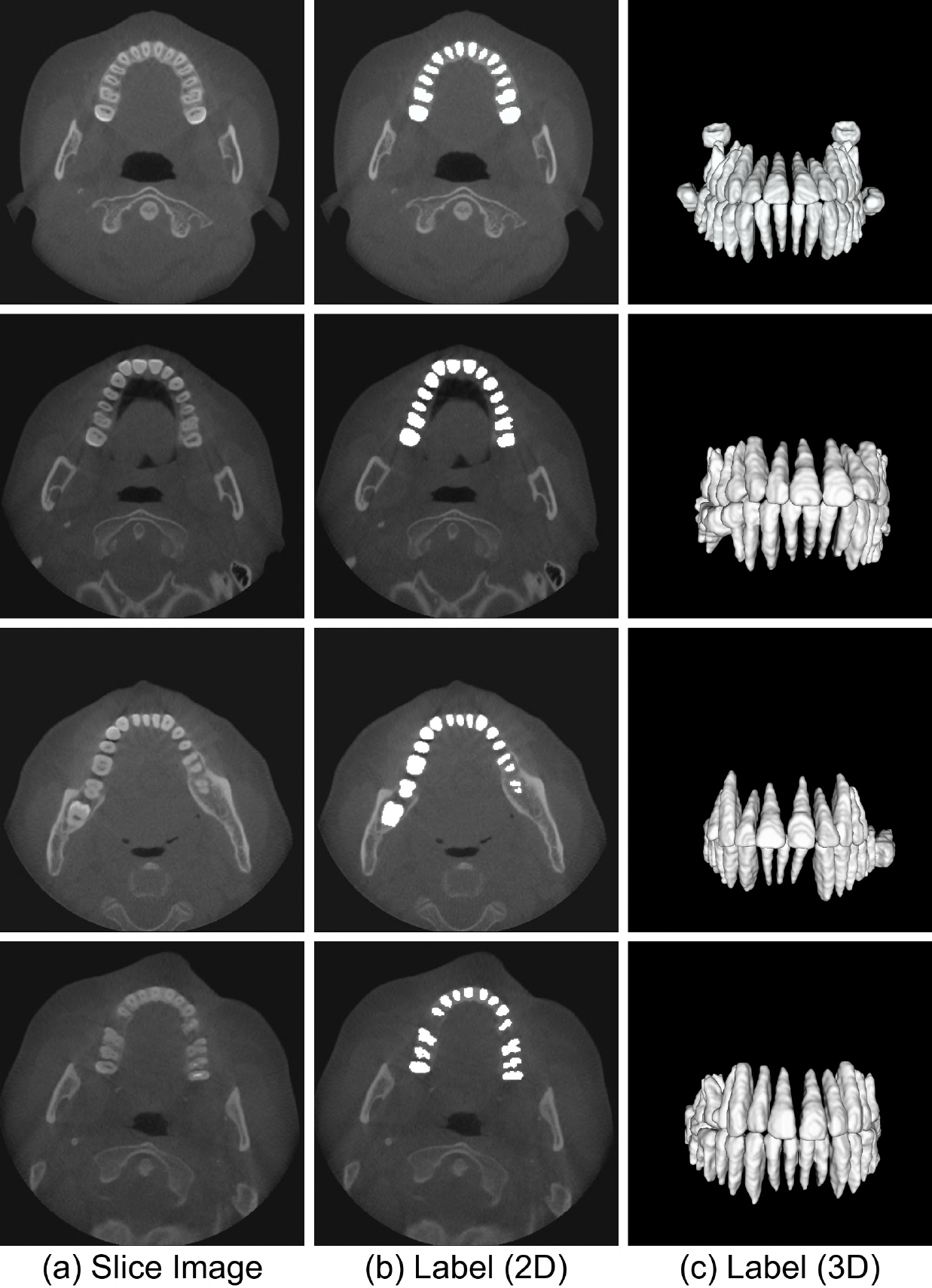}
\caption{\label{fig:3d dataset}The 3D CBCT tooth dataset samples.}
\end{figure}

\textbf{3D CBCT Dataset.} The 3D CBCT dataset used in our study is collected from a large-scale CBCT dataset released by \cite{cui2022fully}. This large-scale dataset is  used for segmentation and reconstruction of individual teeth and alveolar bone, and it consists of 4938 CBCT scans from 15 different centers in China with varying data distributions. 
However, only partial data has been released due to the privacy issues and regulation polices in hospitals. The data setting in our study is identical with \cite{PMFSNet}, which 129 scans are used in total, dividing into a training set of 103 scans and a test set of 26 scans. The physical resolution of these scans is isotropic resolution, varying from 0.2 to 0.4 $mm^3$. Some samples are showed in Figure \ref{fig:3d dataset}.

\noindent
\textbf{2D TED3 Dataset.} Our proposed TED3 dataset is also used for the evaluation of our framework T-Mamba. The detailed description of this dataset has been mentioned in Sec. \ref{sec3:ted3}.

\subsubsection{Implementation Details}

\textbf{Experiments on 3D CBCT data.} The image pre-processing and data augmentation is strictly the same as \cite{PMFSNet}. We resample each 3D image to a uniform voxel spacing of 0.5 × 0.5 × 0.5mm and then randomly crop every image to 160 × 160 × 96 size for model inputs.
The training process incorporates the AdamW optimizer with betas set to (0.9, 0.999), a momentum of 0.8, a ReduceLROnPlateau learning rate scheduler, an initial learning rate of 5e\--{}3, and a weight decay of 5e\--{}5. 
Following the configuration in \cite{PMFSNet}, T-Mamba undergo training from scratch for 20 epochs.
Five evaluation metrics were employed to assess the network’s performance, including Dice Similarity Coefficient (DSC), Intersection over Union (IoU), Mean Intersection over Union (mIoU), Accuracy (ACC), Hausdorff Distance (HD), Average Symmetric Surface Distance (ASSD), and Surface Overlap (SO).

\noindent
\textbf{Experiments on 2D X-ray data.} 
The color jitterring and image rotation are utilized as data augmentations in the training stage. The image is resized to 640 × 1280 before feeding to the network. The batch size is 16 and the AdamW optimizer is employed . The T-Mamba is trained for a total of 30 epochs with an initial learning rate of 0.0075 and weight decay of 0.00001, and the MultiStepLR strategy is used to tune the learning rate. Three metrics were employed to assess the model's performance including Dice Similarity Coefficient (DSC), Intersection over Union (IoU), and Accuracy (ACC). All experiments are conducted using pyTorch on four NVIDIA V100 GPUs. 

\subsection{Quantitative and Qualitative Comparisons}

\subsubsection{Comparisons with the state-of-the-arts on 3D tooth CBCT}

\begin{table*}[h]\footnotesize
\centering
\caption{Comparison results of different methods on the 3D CBCT tooth dataset. The best results are in bold. $\uparrow$ means higher values are better, $\downarrow$ means lower values are better. Results for most methods are taken from PMFSNet3D.}
\label{tab: 3D CBCT Tooth Comparison}
\setlength{\tabcolsep}{0.1mm}{
    \begin{tabular}{lccccccc}
    \toprule
    Method & FLOPs(G)$\downarrow$ & Params(M)$\downarrow$ & HD(mm)$\downarrow$ & ASSD(mm)$\downarrow$ & IoU(\%)$\uparrow$ & SO(\%)$\uparrow$ & DSC(\%)$\uparrow$ \\
    \midrule
    UNet3D \cite{UNet3D} & 2223.03 & 16.32 & 113.79 & 22.40 & 70.62 & 70.72 & 36.67 \\
    DenseVNet \cite{DenseVNet} & 23.73 & 0.87 & 8.21 & 1.14 & 84.57 & 94.88 & 91.15 \\
    AttentionUNet3D \cite{AttentionUNet} & 2720.79 & 94.48 & 147.10 & 61.10 & 52.52 & 42.49 & 64.08 \\
    DenseVoxelNet \cite{DenseVoxelNet} & 402.32 & 1.78 & 41.18 & 3.88 & 81.51 & 92.50 & 89.58 \\
    MultiResUNet3D \cite{MultiResUNet3D} & 1505.38 & 17.93 & 74.06 & 8.17 & 76.19 & 81.70 & 65.45 \\
    UNETR \cite{UNETR} & 229.19 & 93.08 & 107.89 & 17.95 & 74.30 & 73.14 & 81.84 \\
    SwinUNETR \cite{SwinUNETR} & 912.35 & 62.19 & 82.71 & 7.50 & 83.10 & 86.80 & 89.74 \\
    TransBTS \cite{TransBTS} & 306.80 & 33.15 & 29.03 & 4.10 & 82.94 & 90.68 & 39.32 \\
    nnFormer \cite{nnFormer} & 583.49 & 149.25 & 51.28 & 5.08 & 83.54 & 90.89 & 90.66 \\
    3D UX-Net \cite{3DUXNet} & 1754.79 & 53.01 & 108.52 & 19.69 & 75.40 & 73.48 & 84.89 \\
    PMFSNet3D \cite{PMFSNet} & 15.14 & 0.63 & 5.57 & 0.79 & 84.68 & {95.10} & {91.30} \\
    SegMamba  & 2254.30 & 67.36  & 3.95 & 0.6347 & 86.60 & 98.07 & {92.75} \\
    
    \midrule
    \textbf{T-Mamba (Ours)} & \textbf{25.43} & \textbf{1.04} & \textbf{1.18} & \textbf{0.42} & \textbf{88.31} & \textbf{97.53} & \textbf{93.60} \\
     & \textbf{-} & \textbf{-} & \textbf{-4.39} & \textbf{-0.37} & \textbf{+3.63} & \textbf{+2.43} & \textbf{+2.30} \\
    \bottomrule
    \end{tabular}}
\end{table*}

\begin{figure}[h]
\centering
\includegraphics[width=1\linewidth]{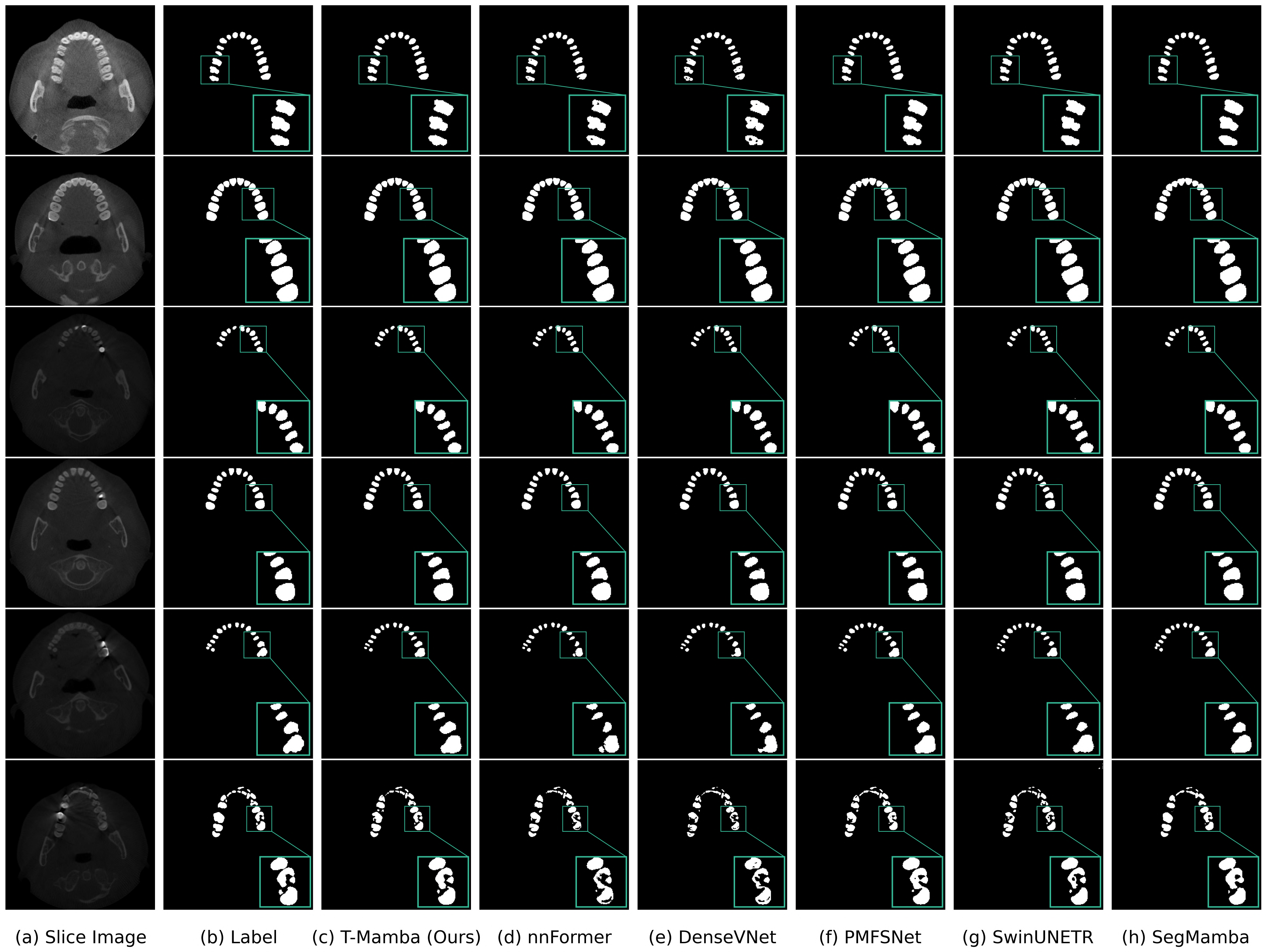}
\caption{Visual Evaluation of T-Mamba Against State-of-the-Art Methods on 3D CBCT tooth}
\label{fig:3d_sample_1}
\end{figure}

\begin{figure}[h]
\centering
\includegraphics[width=1\linewidth]{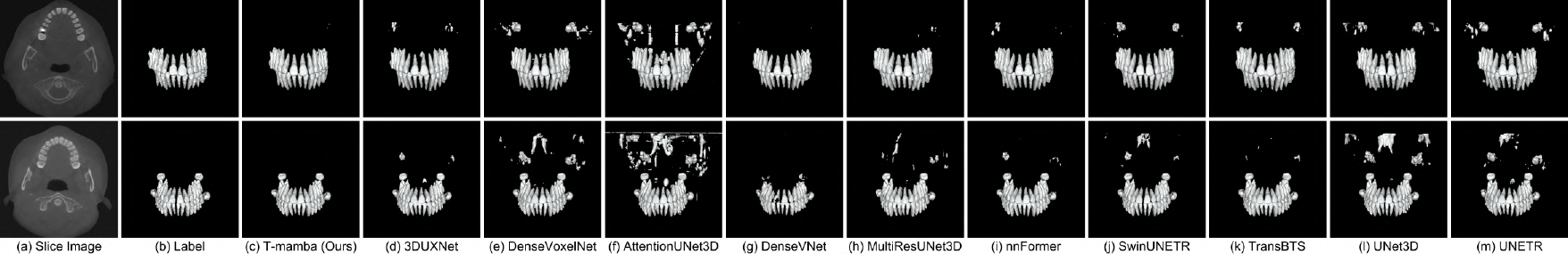}
\caption{Visual Evaluation of T-Mamba Against State-of-the-Art Methods on 3D CBCT tooth}
\label{fig:3d_compair_vis_3d}
\end{figure}

To evaluate the performance of our T-Mamba on the 3D tooth CBCT data, we comprehensively compare it with several state-of-the-art (SOTA) 3D networks spanning a range of neural architectures. These networks includes the UNet3D and its variants (MultiResUNet3D, AttentionUNet3D and PMFSNet3D), the DenseVNet and its variants (DenseVoxelNet), the transformer-based networks like UNETR, SwinUNETR, TransBTS, nnFormer, as well as 3D UX-Net, and vision mamba architecture like SegMamba. Table \ref{tab: 3D CBCT Tooth Comparison} demonstrates the comparisons in terms of the accuracy and computational complexity. Overall, T-Mamba achieves the best results in all metrics compared to current state-of-the-art methods, and it outperforms the previous SOTA method PMFSNet3D from accuracy perspectives by a large margin. More specifically, T-Mamba reduces the Hausdorff Distance (HD) by \textbf{4.39 mm} and the Average Symmetric Surface Distance (ASSD) by \textbf{0.37 mm}. Additionally, T-Mamba enhances the Intersection over Union (IoU) by \textbf{3.63\%}, the Similarity Overlap (SO) by \textbf{2.43\%}, and the Dice Similarity Coefficient (DSC) by \textbf{2.30\%}.
T-Mamba's superior performance can be attributed to its innovative integration of shared dual positional encoding and frequency-based features into the vision mamba architecture, along with a gate selection unit that adaptively combines spatial and frequency domain features, enabling it to better handle long-range dependencies and enhance feature representation in 3D CBCT tooth segmentation tasks.
Figure \ref{fig:3d_sample_1} presents a qualitative comparison on a variety of teeth CBCT slices. Green boxes are used to highlight specific areas of interest, particularly regions where segmentation differences between models are more pronounced. It is evident that T-Mamba excels particularly well in handling more complex cases compared to other methods, especially in scenarios involving significant structural anomalies. Figure \ref{fig:3d_compair_vis_3d} shows a qualitative comparison on 3D tooth CBCT data segmentation. 

\subsubsection{Comparisons with the state-of-the-arts on TED3 dataset}

\begin{table*}[h]\footnotesize
\centering
\caption{Comparison results of different methods on the TED3 dataset}
\label{tab: 2D TED3 Tooth Comparison}
\setlength{\tabcolsep}{0.7mm}{
    \begin{tabular}{lccccccc}
    \toprule
    Method & IoU(\%)$\uparrow$ & DSC(\%)$\uparrow$ & ACC(\%)$\uparrow$ & Params(M)$\downarrow$ \\
    \midrule
    UNet  & 80.67 & 87.89 & 96.42 & 31.04   \\
    Attention-UNet  & 80.63 & 87.66 & 96.43 & 34.88   \\
    BCDU-Net  & 80.04 & 87.42 & 96.22 & 18.45   \\
    CE-Net  & 81.00 & 87.92 & 96.42 & 29.00   \\
    CPF-Net  & 81.51 & 88.30 & 96.52 & 43.27   \\
    CKD-Net  & 77.52 & 84.97 & 95.65 & 59.34   \\
    PMFSNet2D  & 81.31 & 88.15 & 96.51 & 0.99   \\
    MaskDINO  & 80.41 & 87.49 & 96.29 & 52.0   \\
    GEM  & 81.43 & 88.06 & 96.42 & 21.6   \\
    SwinUMamba  & 79.78 & 87.32 & 96.29 & 59.88   \\
    \midrule
    Ours  & 81.52 & 88.25 & 96.55 & 0.68   \\
    Ours + pseudo-label  & 82.41 & 88.77 & 96.78 & 0.68   \\
    \bottomrule
    \end{tabular}}
\end{table*}

We also evaluate the performance of our T-Mamba on the TED3 dataset by comparing it with several state-of-the-art (SOTA) 2D networks designed for tooth segmentation on panoramic radiographs. 
These networks encompass a variety of architectures, including UNet and its variants (Attention-UNet, BCDU-Net), CNN-based models (CE-Net, CPF-Net, CKD-Net, BiseNetv2, PMFSNet2D), transformer-based models (MaskDINO, GEM), and SwinUMamba. Table 3 demonstrates the comparisons in terms of accuracy and model complexity. 
It can be shown that our T-Mamba consistently achieves the highest accuracy across all metrics on the TED3 dataset (\textbf{81.52\%} IoU, \textbf{88.25\%} DSC, \textbf{96.55\%} ACC), and the performance could be further enhanced by introducing the unlabelled data. Notably, our approach achieves these superior results while maintaining significantly fewer parameters. Specifically, our model utilizes only 0.68M parameters, substantially fewer than most competing models.
These findings underscore that our method not only achieves state-of-the-art accuracy on the TED3 dataset but does so with exceptional efficiency, highlighting its potential for practical applications in dental image analysis.

\begin{figure}[h!]
\centering
\includegraphics[width=1\linewidth]{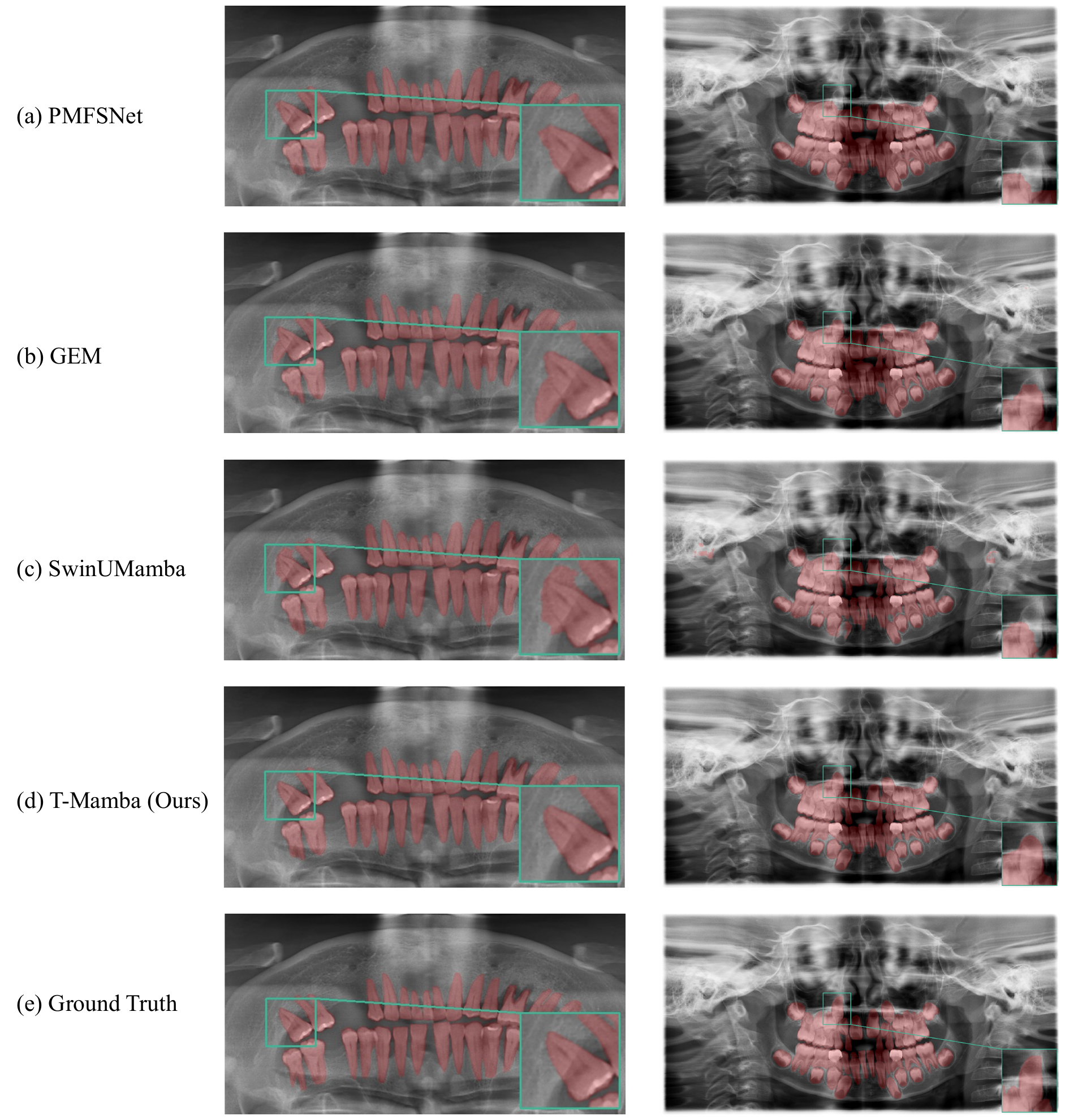}
\caption{Qualitative Evaluation of T-Mamba Against State-of-the-Art Methods on typical panoramic images on TED3 test set.}
\label{fig:2d_sample_1}
\end{figure}

\begin{figure}[h!]
\centering
\includegraphics[width=1\linewidth]{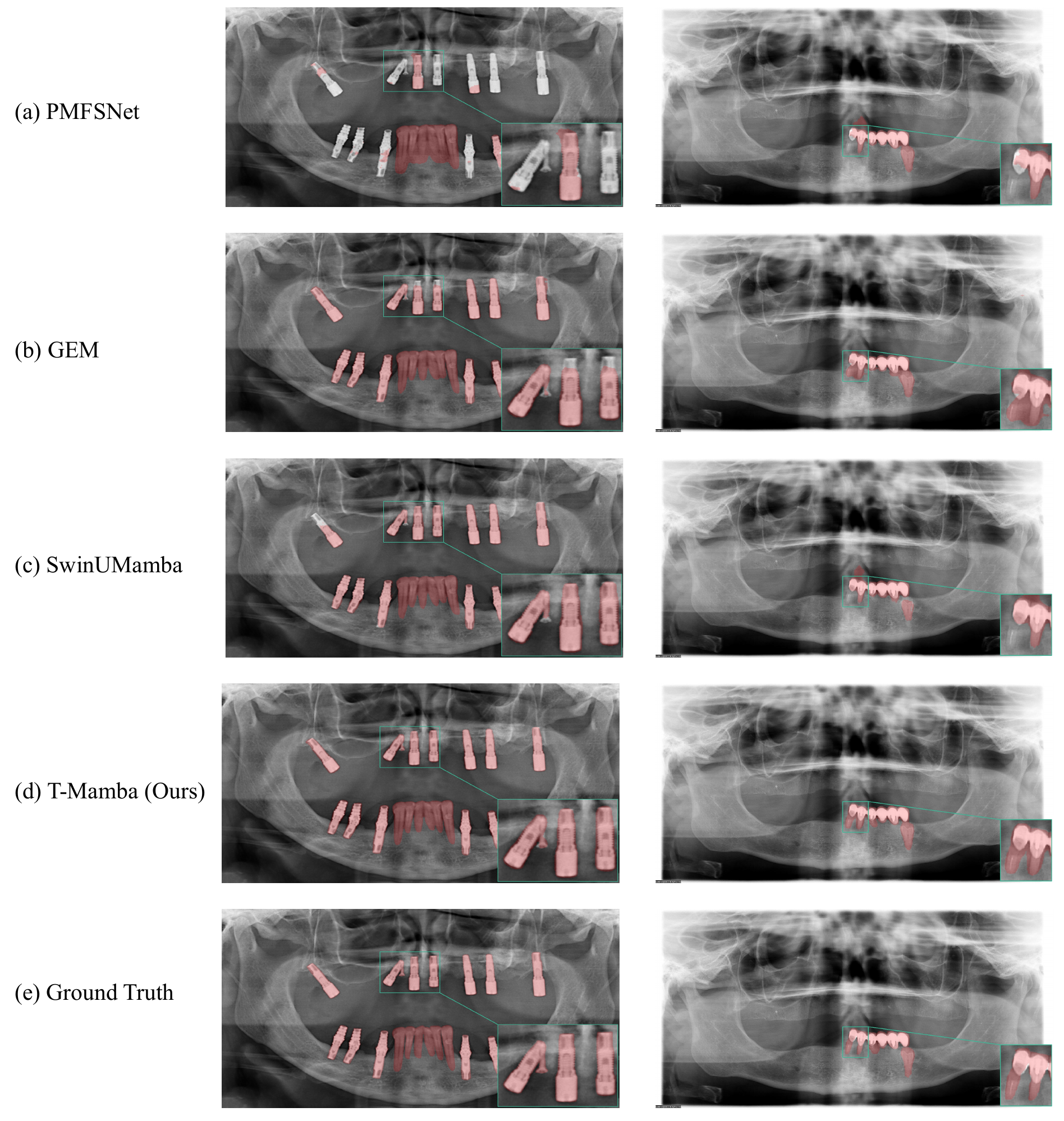}
\caption{Qualitative Evaluation of T-Mamba Against State-of-the-Art Methods on teeth missing cases on TED3 test set.}
\label{fig:2d_sample_2}
\end{figure}

Figure \ref{fig:2d_sample_1} and Figure \ref{fig:2d_sample_2} showcase a comparison of tooth segmentation results from the test set in TED3 dataset using different state-of-the-art models. These samples present typical panoramic dental X-ray images showing a full view of the upper and lower jaws, and some more challenging case with multiple missing teeth. The comparison includes results from (a) PMFSNet, (b) GEM, (c) SwinUMamba, (d) T-Mamba (our proposed method), and (e) Ground Truth. The segmentation results are overlaid in red on the original grayscale X-ray images. All methods perform relatively well on normal cases, with T-Mamba providing the segmentation most similar to the Ground Truth, especially in capturing the detailed contours of individual teeth and maintaining consistency across the entire dental arch. Regarding the teeth missing cases, various methods show different levels of accuracy, with T-Mamba being the most accurate one in segmenting the remaining teeth and properly identifying the toothless areas.

\subsection{Ablation study}
\subsubsection{The architecture components.}
We conduct experiments on 3D CBCT dataset to show the effectiveness of our proposed components in Tim block by adding them one at a time, which is shown in Table \ref{tab: ablation study of Tim block}. 
First and foremost, we directly add vanilla Vim block into DenseVNet for introducing the global feature relationship modeling with linear complexity. The metric IoU, SO, DSC improve by 2.79\%, 0.93\%, 1.89\%, respectively; but the metric HD and ASSD get worse by 1.32 and 0.11mm, respectively. This phenomenon indicates that the Vim block which could capture the long-range dependency from image features is beneficial to tooth CBCT segmentation while still exists some shortcomings. 
To further enhance the performance of Vim block, we introduce three components: 1) Shared bi-positional encoding compensation; 2) Frequency-based bandpass filtering; 3) Gated selection unit.

Regarding the positional encoding compensation, all metrics except HD get worse when only using the pre-position embedding which is inserted before norm operation in Tim block. On the other hand, only the metric ASSD and SO are improved when utilizing the post-position embedding which is inserted in gate selection unit. To our surprised, the shared bi-position embedding could bring significant improvements in all metrics, especially for HD and ASSD metrics. These three experiments show that the shared bi-position encoding is crucial for compensating the loss of positional information during the reshape operations of input and output, and can improve segmentation accuracy without introducing extra model parameters. 
Afterwards, we add the residual connection from input sequence to output sequence, and this step can bring slight increases in all metrics. 
We also leverage the frequency-based bandpass filtering to extract the distinctive feature representation in the frequency domain. There is a small rise in the metric IoU and DSC. 
Ultimately, the gate selection unit is added to adaptively fuse three kinds of features and this unit dramatically increase the metric IoU, SO, DSC. 
To sum up, our proposed Tim block enhances the metrics IoU, SO, DSC by 3.74\%, 2.65\%, 2.45\%, respectively, and reduces the distance HD and ASSD by 7.03mm and 0.72mm, respectively, compared with DenseVNet baseline. Besides, our Tim block achieves higher results compared with Vim block. 

\begin{table*}[t]\small
\centering
\caption{The ablation study of our Tim block. $\uparrow$ means higher values are better, $\downarrow$ means lower values are better.}
\label{tab: ablation study of Tim block}
\setlength{\tabcolsep}{0.1mm}{
    \begin{tabular}{lccccccc}
    \toprule
    Method & HD(mm)$\downarrow$ & ASSD(mm)$\downarrow$ & IoU(\%)$\uparrow$ & SO(\%)$\uparrow$ & DSC(\%)$\uparrow$ \\
    \midrule
    DenseVNet & 8.21 & 1.14 & 84.57 & 94.88 & 91.15 \\
    \textbf{+ Vim} &9.53	&1.25 &87.36 &95.81	&93.04 \\
    \midrule
    + Pre Pos & 7.89	& 1.79	& 87.00	& 95.35	& 92.80 \\
    + Post Pos & 10.65	& 1.00	& 87.10	& 96.23	& 92.87 \\
    \textbf{+ Shared bi-Pos} & 1.22	& 0.59	& 87.65	& 96.79	& 93.20 \\
    \midrule
    \textbf{+ Residual Connection} & \textbf{1.10}	& \textbf{0.39}	& 87.95	& 97.48	& 93.40 \\
    \midrule
    \textbf{+ FFT Bandpass Filtering} & 1.18	& 0.67	& 88.05	& 97.26	& 93.47 \\
    \midrule
    \textbf{+ Gate Selection Unit} & 1.18	& 0.42	& \textbf{88.31}	& \textbf{97.53}	& \textbf{93.60} \\
    \bottomrule
    \end{tabular}}
\end{table*}

\subsubsection{The threshold of bandpass filtering.}
We conduct extensive experiments to search the best thresholds for the high and low frequency of the bandpass filtering in Tim block, which is shown in Table \ref{tab: ablation for frequency thresholds}.  We find that the metrics IoU, SO, and DSC achieve the highest results when the values of high and low frequency thresholds are 0.9 and 0.1, respectively. In addition, the metrics HD and ASSD get lowest distance when these two thresholds are setting as 0.6 and 0.4, respectively. We select 0.9 and 0.1 thresholds for the high and low frequency as our default settings for other experiments. 

\begin{table*}[h]\footnotesize
\centering
\caption{The ablation experiments on thresholds for the high and low frequency of the bandpass filtering in Tim block. The best results are in bold. $\uparrow$ means higher values are better, $\downarrow$ means lower values are better.}
\label{tab: ablation for frequency thresholds}
\setlength{\tabcolsep}{3mm}{
    \begin{tabular}{cccccccc}
    \toprule
    Low Threshold & High Threshold $\downarrow$ & HD(mm)$\downarrow$ & ASSD(mm)$\downarrow$ & IoU(\%)$\uparrow$ & SO(\%)$\uparrow$ & DSC(\%)$\uparrow$ \\
    \midrule
    0.1 & 0.9  & 1.18 & 0.42 & \textbf{88.31} & \textbf{97.53} & \textbf{93.60} \\
    0.2 & 0.8  & 1.18 & 0.41 & 87.77 & 97.16 & 93.28 \\
    0.3 & 0.7  & 1.20 & 0.53 & {87.81} & {97.14} & {93.31} \\
    0.4 & 0.6  & \textbf{1.16} & \textbf{0.34} & {87.91} & {97.49} & {93.35} \\
    0.5 & 0.5  & 1.25 & 0.50 & {87.71} & {97.12} & {93.324} \\
    \bottomrule
    \end{tabular}}
\end{table*}

\subsection{The effect of Gated Selection Unit.}
\begin{figure}[H]
    \centering
    \begin{subfigure}{0.45\textwidth}
        \centering
        \includegraphics[width=\textwidth]{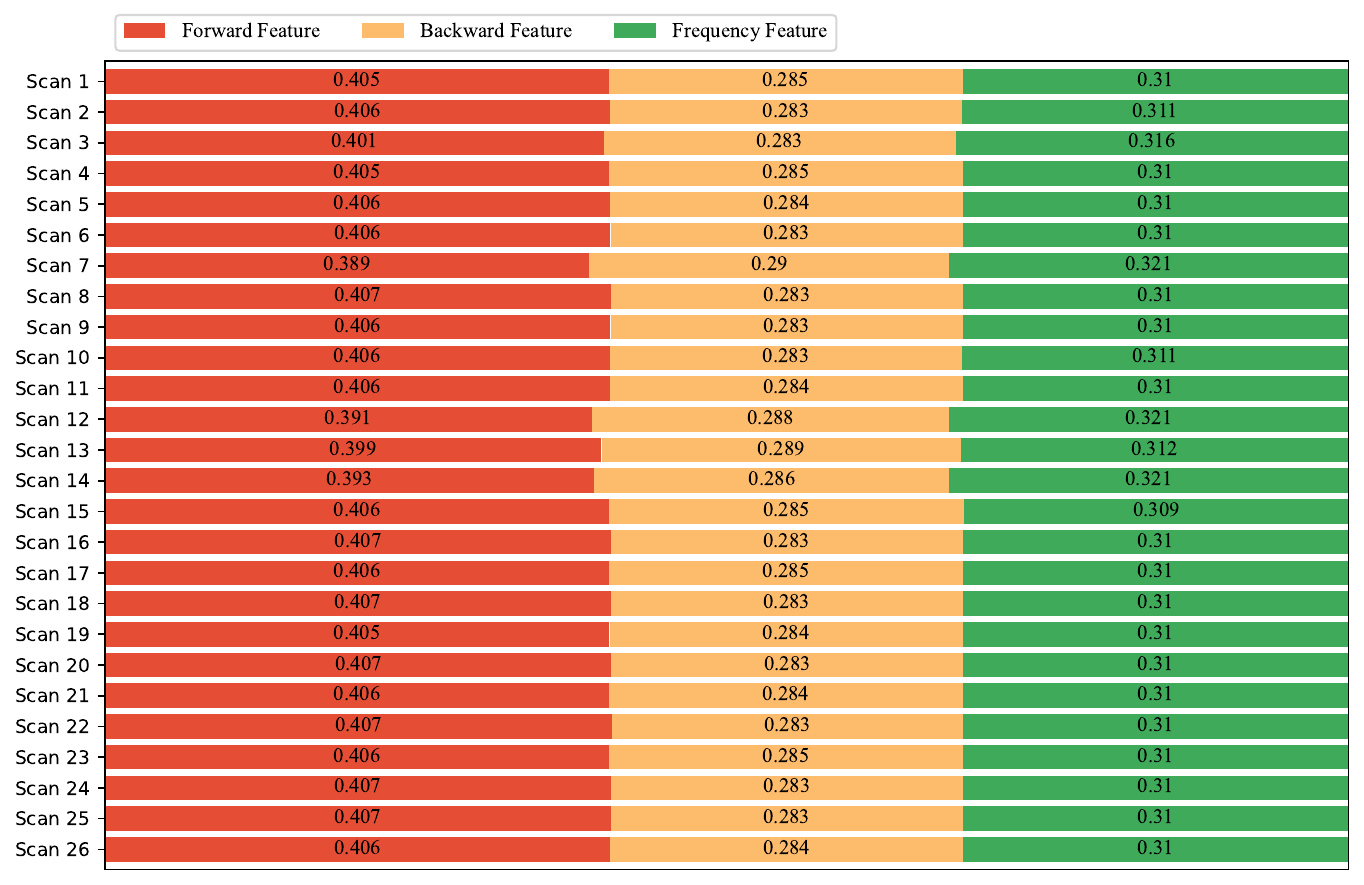}
        \caption{Proportions for scans}
        \label{fig:sub1}
    \end{subfigure}
    \begin{subfigure}{0.45\textwidth}
        \centering
        \includegraphics[width=\textwidth]{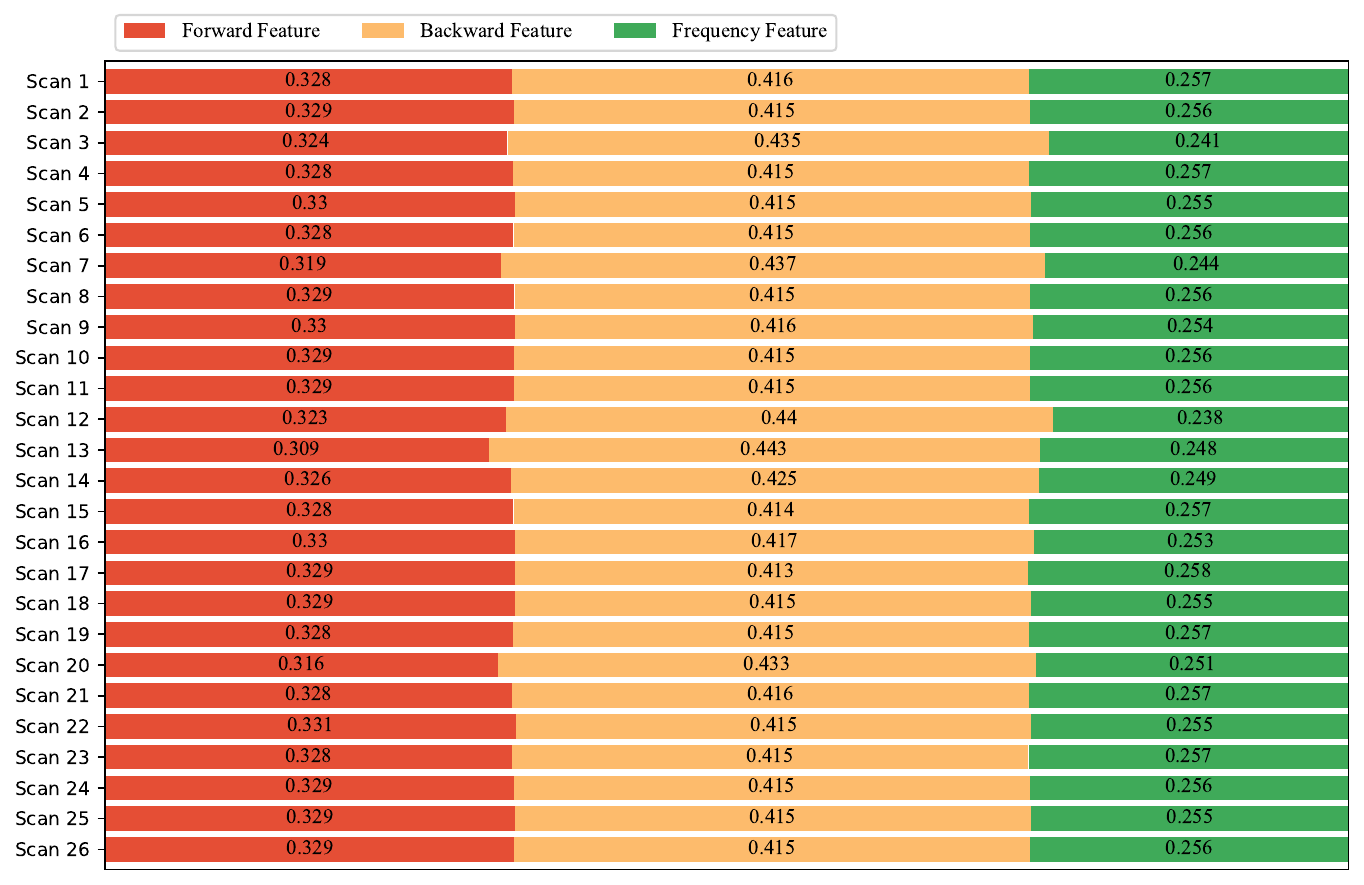}
        \caption{Proportions for 1st feature scales}
        \label{fig:sub2}
    \end{subfigure}


    \begin{subfigure}{0.45\textwidth}
        \centering
        \includegraphics[width=\textwidth]{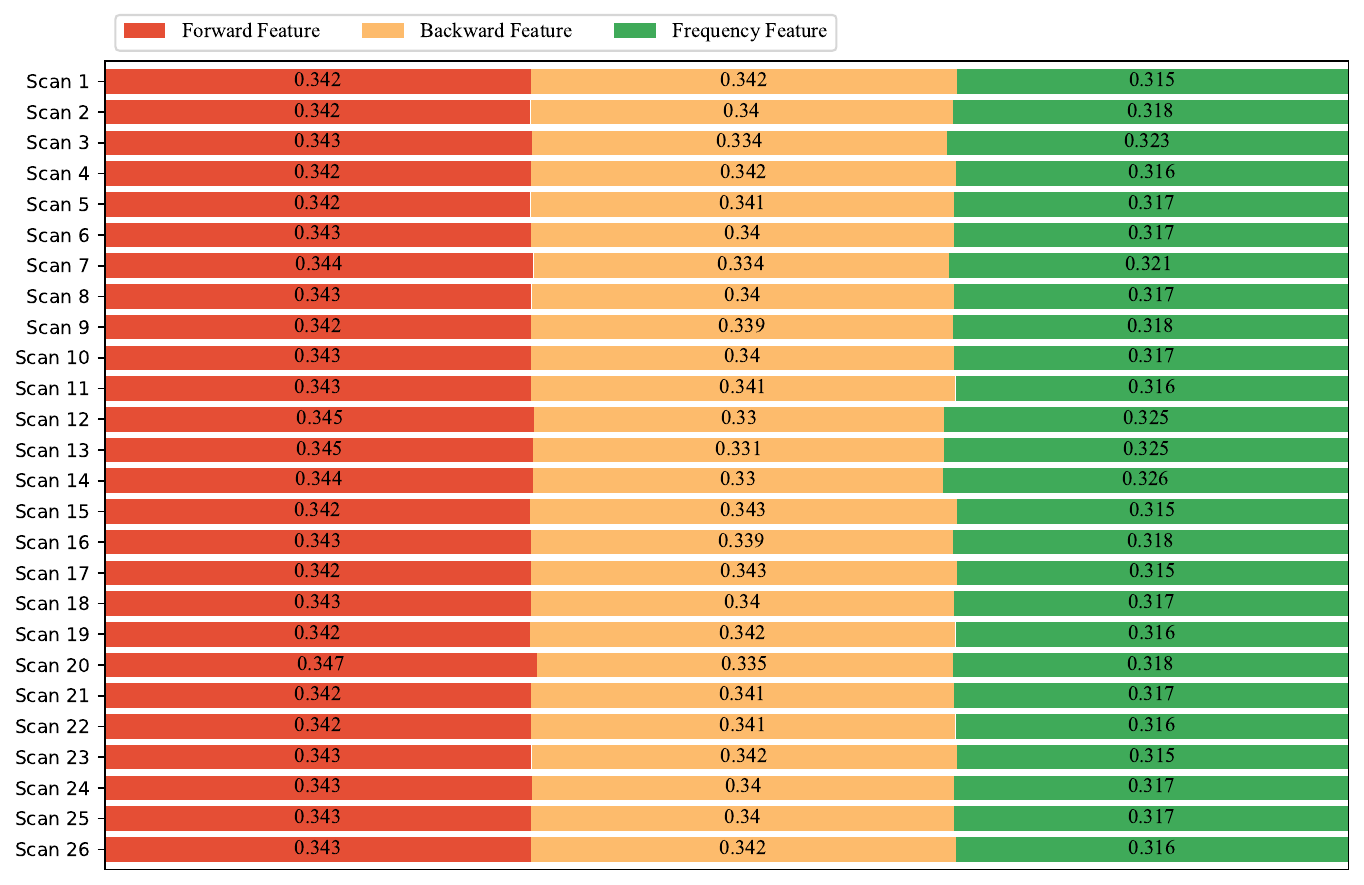}
        \caption{Proportions for 2nd feature scales}
        \label{fig:sub3}
    \end{subfigure}
    \begin{subfigure}{0.45\textwidth}
        \centering
        \includegraphics[width=\textwidth]{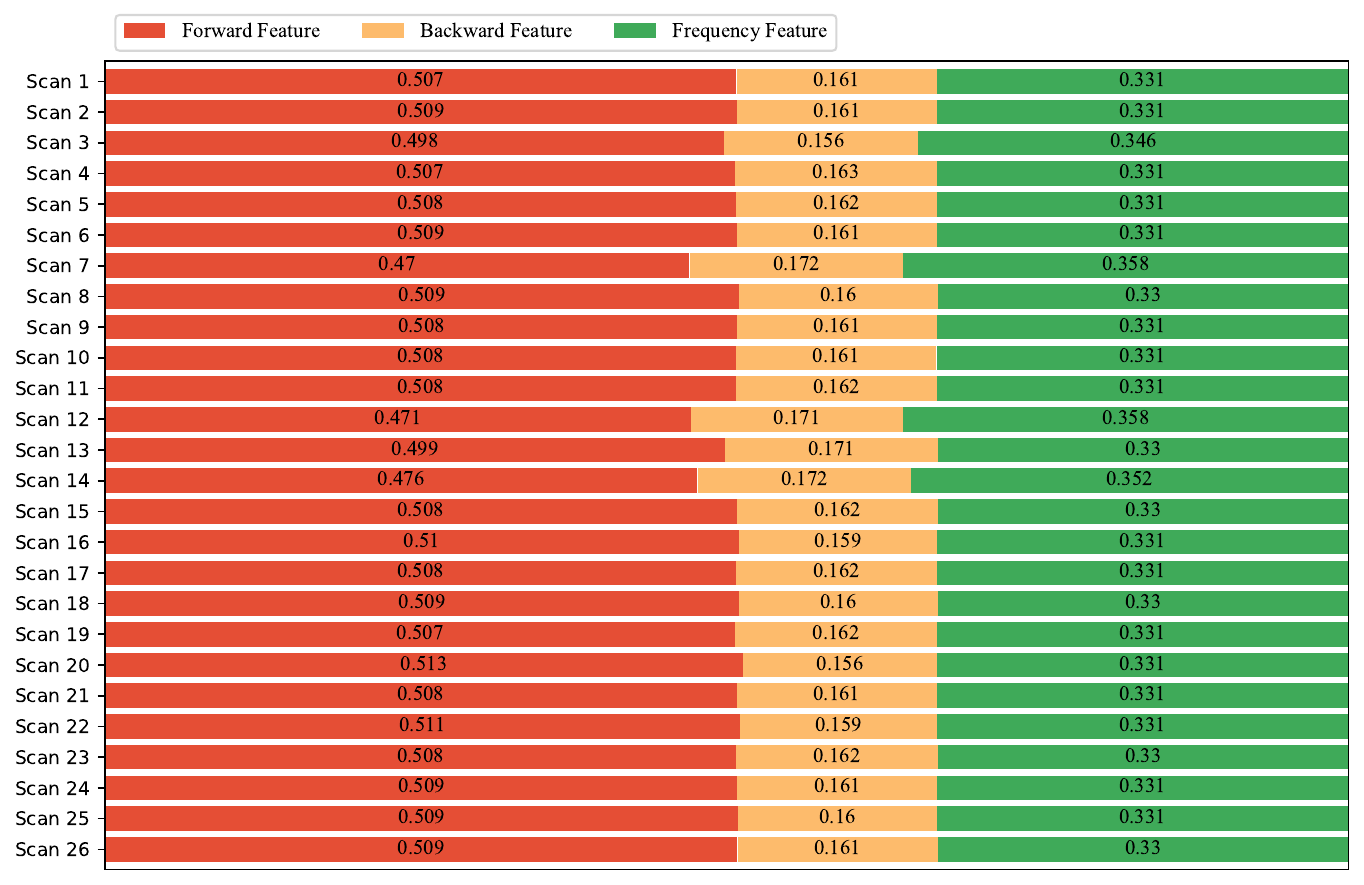}
        \caption{Proportions for 3rd feature scales}
        \label{fig:sub4}
    \end{subfigure}
    \caption{Four subfigures in two rows and two columns}
    \label{fig: GSU all stages}
\end{figure}

The Gated Selection Unit in Tim block is to adaptively fuse two features in spatial domain (both forward and backward directions) and one feature in frequency domain. We visualize the average proportions of these three different features on three feature-scales shown as in Figure \ref{fig: GSU all stages} (a) and the proportions on three specific feature-scales shown as in Figure \ref{fig: GSU all stages} (b-d) for all validation scans.
In Figure \ref{fig: GSU all stages} (a), we can observe that the forward feature, the backward feature, and frequency feature approximately occupy 41\%, 28\%, 31\%, respectively. This could also prove that the features in frequency domain are more essential than backward feature in spatial domain. 
However, the proportions of these three features vary greatly in different feature-scales. For instance, in the first feature scale which belongs to low-level features, the percentage of backward feature is obviously larger than other two features. Nonetheless, the percentage of forward feature is the highest in the third feature scale which belongs to high-level features, and the the proportions of of these three features are nearly equivalent in the second feature scale.

Hence, we can draw the conclusion that our proposed Gated Selection Unit has the ability to integrate two features in spatial domain and and one feature in frequency domain adaptively, thus the model can extract more fine-grained and robust features.

\subsection{Mamba v.s. Self-Attention.}

T-Mamba endows the DenseVNet with the ability of global dependency modeling through the Tim module. We compared the effects of the vanilla self-attention module and our proposed Tim module directly added to the DenseVNet on the tooth CBCT dataset. As shown in Table \ref{tab: mamba self attn}, our Tim module greatly exceeds the popular self-attention module in terms of accuracy and inference speed, proving the effectiveness and efficiency of the Tim module in extracting long-range dependency of medical images.

\begin{table*}[h]\footnotesize
\centering
\caption{The ablation experiments on two global dependency modeling operations, Mamba and Self-Attention. ``Time'' refer to the inference time per CBCT data. The best results are in bold. $\uparrow$ means higher values are better, $\downarrow$ means lower values are better.}
\label{tab: mamba self attn}
\setlength{\tabcolsep}{2mm}{
    \begin{tabular}{lccccccc}
    \toprule
    Method & HD(mm)$\downarrow$ & ASSD(mm)$\downarrow$ & IoU(\%)$\uparrow$ & SO(\%)$\uparrow$ & DSC(\%)$\uparrow$ & Time (min) $\downarrow$\\
    \midrule
     DenseVNet + Self-Attn & 48.04	&6.78 &84.07 &83.79	&90.08 & 3.96\\
    \textbf{DenseVNet + Tim (ours)}  & \textbf{1.18}	& \textbf{0.42}	& \textbf{88.31}	& \textbf{97.53}	& \textbf{93.60} & \textbf{2.42}\\
    \bottomrule
    \end{tabular}}
\end{table*}

\subsection{Visualization of the Grad-CAM.}

\begin{figure}[h]
\centering
\includegraphics[width=0.7\linewidth]{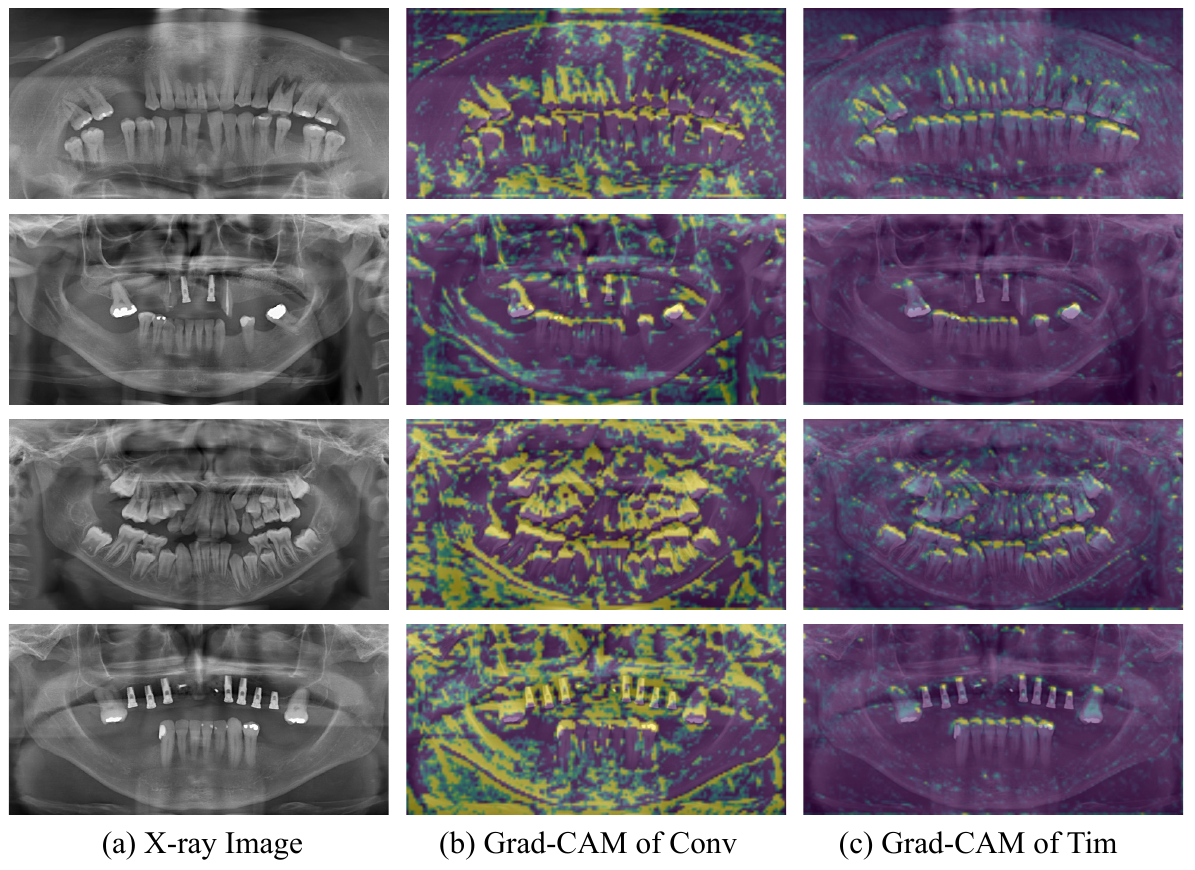}
\caption{Grad-CAM visualizations on four dental X-ray images. The first column (a) shows the original X-ray images, the second column (b) presents the Grad-CAM visualizations generated by a convolution operation, and the third column (c) shows the Grad-CAM visualizations using our Tim module.}
\label{fig:heatmap}
\end{figure}

We used the Grad-CAM \cite{selvaraju2017grad} visualization approach to compare the feature map before and after the Tim operation, and it can be observed in Figure \ref{fig:heatmap}.
Our Tim module assigns higher weights to the area of the tooth thanks to the strength of the global dependency modeling inherited in Tim module, highlighting the regions in the image that are most important for the model’s predictions.

\section{Conclusion}
We have proposed T-Mamba to achieve global and local visual context modeling for tooth segmentation. Thanks to the proposed Tim block that integrates shared positional encoding and frequency-based features into vision mamba, we address limitations in spatial position preservation and feature enhancement in frequency domain for medical images which are of high noise and low contrast. 
Extensive experiments demonstrate that T-Mamba achieves new SOTA results on the public tooth CBCT dataset, showing that Tim has great potential to be the next-generation feature extractor for efficient long-range dependency modeling in biomedical image analysis. We also propose a public teeth large-scale 2D X-Ray dental dataset, TED3, and our model outperforms previous SOTA methods in a large margin on TED3 dataset. Hoping this dataset can be served as a valuable asset for propelling the application of artificial intelligence in the field of dentistry

In future works, we endeavor to explore the self-supervised learning, such as mask image modeling pretraining, in conjunction with our proposed Tim block for tooth CBCT and X-ray perception tasks. We anticipate that this advancement in technology will further propel the evolution of modern digital dentistry.



\printbibliography

\clearpage
\appendix

\section{Dataset Source}
\label{appendix1}

\begin{table*}[h]\footnotesize
\centering
\caption{The list of dataset source contained in our TED3 dataset.}
\label{tab: data sources}
\setlength{\tabcolsep}{1mm}{
    \begin{tabular}{cccccc}
    \toprule
    No. & Data Modality & Number & Data description & Source \\
    \midrule
    1 & 2D X-ray & 7000 & \makecell[c]{Tooth Semantic Segmentation \\ 
    3500 labelled masks \& 3500 unlabelled data } & \cite{dataset_1} \\
    \hline
    2 & 2D X-ray & 116 & \makecell[c]{Mandible Semantic Segmentation \\ 116 labelled masks} & \cite{dataset_2} \\
    \hline
    3 & 2D X-ray & 1272 & \makecell[c]{Abnormal Tooth Detection \\ Classes: `Cavity', `Fillings', `Impacted Tooth', `Implant'\\1272 labelled boxes } & \cite{dataset_3} \\
    \hline
    4 & 2D X-ray & 598 & \makecell[c]{Tooth Instance Segmentation \\ 598 labelled masks} & \cite{dataset_4} \\
    \hline
    5 & 2D X-ray & 100 & \makecell[c]{Dental caries Semantic Segmentation \\ 100 labelled masks} & \cite{dataset_5} \\
    \hline
    6 & 2D X-ray & 2576 & \makecell[c]{Abnormal Tooth Detection \\ Classes: `caries', `deep caries', `periapical lesions', `impacted teeth' \\1005 labelled boxes \& 1571 unlabelled data} & \cite{dataset_6} \\
    \hline
    7 & 2D X-ray & 543 & \makecell[c]{Tooth Instance Segmentation \\ 543 labelled masks} & \cite{dataset_7} \\
    \hline
    8 & 2D X-ray & 1000 & \makecell[c]{Tooth Instance Segmentation \& Language description \\ 1000 labelled masks} & \cite{dataset_8} \\
    \hline
    9 & 2D X-ray & 1978 & \makecell[c]{Tooth Semantic Segmentation \& Tooth Object Detection \\ 1978 labelled data} & \cite{dataset_9} \\
    \hline
    10 & 2D X-ray & 1988 & \makecell[c]{Tooth Semantic Segmentation \\ 1988 labelled masks} & \cite{dataset_10} \\
    \hline
    11 & 2D X-ray & 4000 & \makecell[c]{Tooth Semantic Segmentation \\ 4000 labelled masks} & \cite{dataset_11} \\
    \hline
    12 & 2D X-ray & 337 & \makecell[c]{panoramic radiography \\ 337 unlabelled data} & \cite{dataset_12} \\
    \hline
    13 & 2D X-ray & 4473 & \makecell[c]{Abnormal Tooth Detection \\ Including 38 classes, 4473 labelled boxes} & \cite{dataset_13} \\
    \hline
    14 & 2D X-ray & 3772 & \makecell[c]{Abnormal Tooth Instance Segmentation \\ Including 9 classes \\ 3772 labelled masks} & \cite{dataset_14} \\
    \hline
    15 & 2D X-ray & 125 & \makecell[c]{Tooth Object Detection \\ 125 labelled boxes} & \cite{dataset_15} \\
    \hline
    16 & 2D X-ray & 10 & \makecell[c]{Tooth Semantic Detection \\ 10 labelled masks} & \cite{dataset_16} \\
    \hline
    17 & 2D X-ray & 2500 & \makecell[c]{Tooth Semantic Segmentation \\ 2000 labelled masks \& 500 unlabelled data} & \cite{dataset_17} \\
    \hline
    18 & 2D X-ray & 3000 & \makecell[c]{Tooth Semantic Detection \\ 3000 labelled masks} & \cite{dataset_18} \\
    \bottomrule
    \end{tabular}}
\end{table*}

\end{document}